\documentclass[letterpaper, 10 pt, conference]{ieeeconf}  

\IEEEoverridecommandlockouts                              

\usepackage{graphics} 
\usepackage{epsfig} 
\usepackage{mathptmx} 
\usepackage{times} 
\usepackage{amsmath} 
\usepackage{amssymb}  
\usepackage{url}
\usepackage{booktabs} 
\usepackage{xcolor} 
\title{\LARGE \bf
Design, Modelling and Characterisation of a Miniature Fibre-Reinforced Soft Bending Actuator for Endoluminal Interventions
}

\author{Xiangyi Tan$^{1}$, Aoife McDonald-Bowyer$^{1}$, Danail Stoyanov$^{1}$ and Agostino Stilli$^{1}$
\thanks{$^{1}$All authors are with the UCL Hawkes Institute, University College London, WC1E 6BT, UK. This work was supported in whole, or in part, by the Wellcome/EPSRC Centre for Interventional and Surgical Sciences (WEISS) [203145/Z/16/Z], the Department of Science, Innovation and Technology (DSIT) and the Royal Academy of Engineering Chair in Emerging Technologies Scheme [CiET 1819/2/36; and the Engineering and Physical Sciences Research Council [EP/ W524335/1]. For the purpose of open access, the author has applied a CC BY public copyright licence to any author accepted manuscript version arising from this submission.;
]. For the purpose of
open access, the authors have applied a CC BY public copyright
licence to any author accepted manuscript version arising from
this submission.
        {\tt\small xiangyi.tan.23@ucl.ac.uk, aoife.mcdonald-bowyer.19@ucl.ac.uk,
        danail.stoyanov@ucl.ac.uk, a.stilli@ucl.ac.uk}}%
}

\begin{document}

\maketitle
\thispagestyle{empty}
\pagestyle{empty}

\begin{abstract}

Miniaturised soft pneumatic actuators are crucial for robotic intervention within highly constrained anatomical pathways. This work presents the design and validation of a fibre-reinforced soft actuator at the centimetre scale for integration into an endoluminal robotic platform for natural-orifice interventional and diagnostic applications. A single-chamber geometry reinforced with embedded Kevlar fibre was designed to maximise curvature while preserving sealing integrity, fabricated using a multi-stage multi-stiffness silicone casting process, and validated against a high-fidelity Abaqus FEM using experimentally parametrised hyperelastic material models and embedded beam reinforcement. The semi-cylindrical actuator has an outer diameter of 18,mm and a length of 37.5,mm. Single and double helix winding configurations, fibre pitch, and fibre density were investigated. The optimal 100 SH configuration achieved a bending angle of 202.9° experimentally and 297.6° in simulation, with structural robustness maintained up to 100,kPa and radial expansion effectively constrained by the fibre reinforcement. Workspace evaluation confirmed suitability for integration into the target device envelope, demonstrating that fibre-reinforcement strategies can be effectively translated to the centimetre regime while retaining actuator performance.

\end{abstract}

\section{INTRODUCTION}

Soft robots are ideally suited to minimally invasive interventions owing to their highly compliant and deformable structures, which allow non-traumatic interaction with surrounding tissues \cite{Runciman2019,Endorobotics2022}. Miniaturised fibre-reinforced bending actuators enable large deformations within highly constrained environments and are increasingly sought in applications where conventional rigid-link mechanisms are unsuitable, particularly in soft interventional and diagnostic systems~\cite{Shi2022RoboSoft,Li2023RAL}. In endoluminal and natural-orifice interventions, such as transoral bronchoscopy, colorectal navigation, upper gastrointestinal screening and trasnvsaginalo access, the device diameter is restricted by anatomy, typically requiring centimetre architectures to prevent trauma during navigation~\cite{McCandless2022,Garbin2018,Arezzo2017,Renda2014}. Natural orifice access enables atraumatic entry without external incisions, but this stringent geometric envelope constrains actuator design space \cite{Kim2022}.

Fibre reinforcement provides directional constraint to convert internal pressure into bending rather than isotropic ballooning~\cite{modeling}. However, when scaled toward the centimetre regime, bending efficiency is often reduced due to tighter minimum winding pitch, increased sensitivity to wall-thickness tolerances, and fabrication-induced material non-uniformity~\cite{Shi2022RoboSoft,mininature, Marechal2020}. This motivates reinforcement strategies that remain effective despite reduced geometric radius and manufacturing tolerance windows.

This paper develops a miniaturised fibre-reinforced bending actuator for soft robotic applications, intended for integration into a novel soft robotic device for endoluminal imaging and intervention. The target anatomy in this paper is the vaginal canal and cervix, a representative endoluminal environment characterised by natural orifice access, a constrained lumen diameter, and a clinical procedure — cervical cell sampling and visual assessment for cancer screening — that demands both precise tip positioning and low-force tissue contact. Using soft robotics in this context may enable a more comfortable and acceptable cervical screening procedure that could address the challenges of declining screening participation \cite{Farajimakin2024BarriersReview, Oscarsson2002WomensStudy} For that purpose, the soft robotic device demonstrated in this paper takes inspiration from the size and dimensions of a tampon. It is cylindrical, with an outer diameter of 18\,mm, a device length of 18\,mm and an actuator body length of 37.5\,mm.
\begin{figure}[t]
  \centering
  \includegraphics[width=0.95\linewidth]{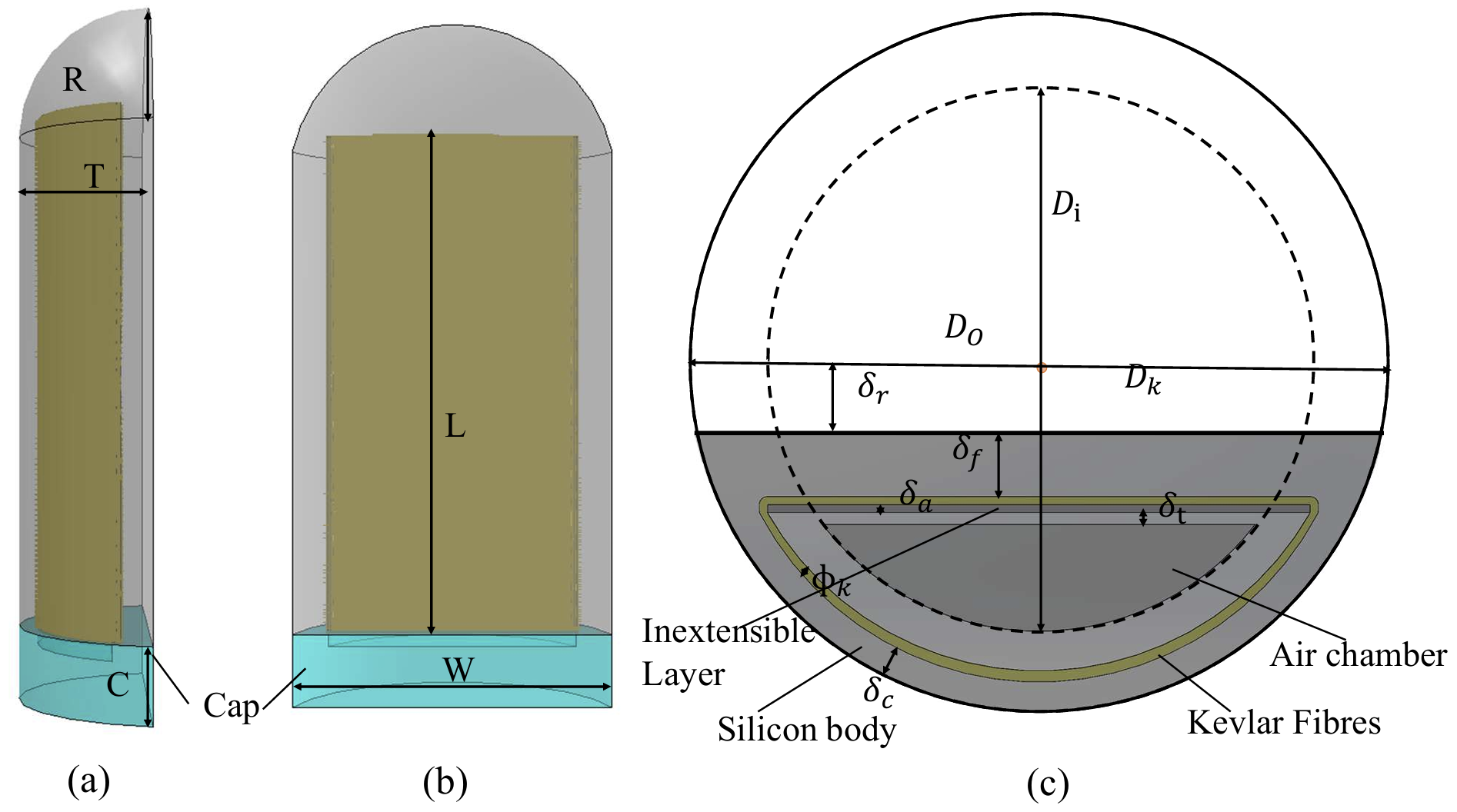}
  \caption{Cross-sectional dimensions and structure of the actuator. The green region denotes the cap, the yellow the Kevlar fibre winding, the thin grey sheet the inextensible fibreglass layer, the dark grey the air chamber, and the semi-transparent grey the silicone body. Symbols correspond to Table~\ref{tab:geomA-params}. (a) Side view, (b) Front view, (c) Cross-sectional view.}
  \label{fig:cross-sections}
\end{figure}
This research contributes to the field in three main ways. First, it adapts fibre-reinforcement principles for a significantly smaller actuator than typically reported in soft robotics literature\cite{Shi2022RoboSoft,Li2023RAL,mininature}, while achieving high curvature at this scale. Second, it provides a high-fidelity FEM modelling framework including explicit reinforcement–matrix coupling, enabling reproducibility~\cite{Chaillou2023,Trivedi2008}. Third, the actuator is evaluated within the context of integration into a miniaturised soft robotic device, linking workspace performance to practical endoluminal navigation constraints.

This paper is organised as follows. Section~II presents the design rationale, fabrication process, and finite element modelling of the miniature fibre-reinforced actuator. Section~III describes the experimental setup and protocols used for validation of the actuator geometry and its integration within the endoluminal device. Section~IV reports the simulation and experimental results. Section~V provides a discussion of the findings and introduces a future extended geometry as a potential enhancement of the current design. Finally, Section~VI concludes the paper and outlines directions for further development toward gynaecological natural-orifice applications.

\section{MATERIAL AND METHODS}
\label{sec:materials-methods}

\subsection{Design of Actuator}
\label{sec:design-actuator-A}
The actuator presented in this paper is shown in Fig. \ref{fig:cross-sections} (a) - (c), with dimensions in \ref{fig:cross-sections} (c) described in Table \ref{tab:geomA-params}. The actuator is designed to form part of a soft robot intended for endoluminal interventional applications, the form factor of which is a cylinder approximately 60 mm in length and 18 mm in diameter. Therefore, the outer diameter of the actuator \(D_O\) is fixed by the device interface and mounting constraints. 
Internal chamber dimensions were selected to maximise curvature in the \(0\!-\!100~\mathrm{kPa}\) pressure range while retaining sufficient silicone cover at the thinnest regions for robustness and demouldability. 
The minimum silicone thickness \(\delta_t\) is \(0.3~\mathrm{mm}\). Chamber depth is \(26.5~\mathrm{mm}\). The cross-sectional chamber area is \(20.8~\mathrm{mm}^2\), corresponding to a nominal chamber volume of approximately \(552~\mathrm{mm}^3\). 

\begin{table}[t]
\caption{GEOMETRICAL PARAMETERS OF THE ACTUATOR}
\label{tab:geomA-params}
\centering
\setlength{\tabcolsep}{3pt}
\renewcommand{\arraystretch}{1.12}
\begin{tabular}{@{}p{1.25cm}p{4.8cm}p{1.6cm}@{}}
\toprule
\textbf{Symbol} & \textbf{Definition} & \textbf{Value} \\
\midrule
\(D_i\) & Internal air chamber diameter & \(14~\mathrm{mm}\) \\
\(D_O\) & Outer geometry diameter & \(18~\mathrm{mm}\) \\
\(\delta_f\) & Silicone cover at flat interface & \(1.6~\mathrm{mm}\) \\
\(\delta_c\) & Silicone cover at curved side & \(0.2~\mathrm{mm}\) \\
\(\delta_t\) & Internal thickness between fibre and chamber & \(0.3~\mathrm{mm}\) \\
\(\delta_r\) & Offset from flat surface to datum origin & \(2~\mathrm{mm}\) \\
\(\delta_a\) & Fibreglass fabric thickness & \(0.2~\mathrm{mm}\) \\
\(\phi_k\) & Kevlar fibre diameter & \(0.206~\mathrm{mm}\) \\
\(D_{\mathrm{rod}}\) & Winding rod diameter & \(5~\mathrm{mm}\) \\
\midrule
\(W\) & Maximum actuator width & \(18~\mathrm{mm}\) \\
\(L\) & Air chamber length & \(26.5~\mathrm{mm}\) \\
\(C\) & Cap length & \(4~\mathrm{mm}\) \\
\(R\) & Length of spherical cap segment & \(7~\mathrm{mm}\) \\
\(T\) & Total actuator thickness & \(7~\mathrm{mm}\) \\
\(L{+}R{+}C\) & Total actuator length & \(37.5~\mathrm{mm}\) \\
\bottomrule
\end{tabular}
\end{table}

\subsection{Fabrication of Actuator}
\begin{figure*}[t]
    \centering
    \includegraphics[width=\textwidth]{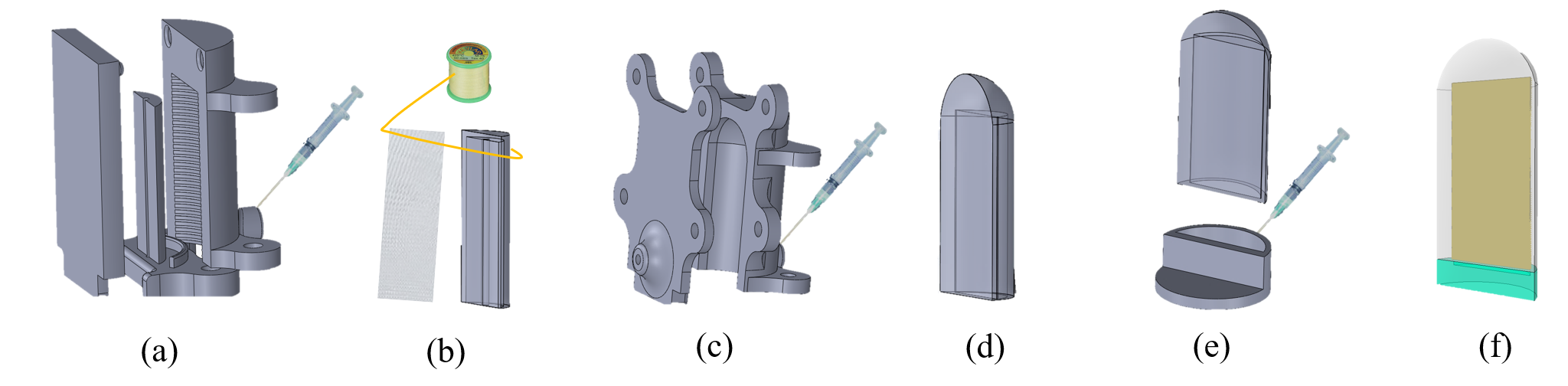}
    \caption{Fabrication of actuator with Geometry~A: (a) Mould assembly for the inner chamber; Ecoflex~00–50 is injected to form the chamber wall. (b) A Kevlar fabric inextensible layer, pre-soaked in Ecoflex~00–50, is bonded to the flat surface of the half-cured inner chamber, and Kevlar fibre are wound through it. (c) The reinforced inner chamber is placed into the outer chamber mould, and Ecoflex~00–50 is injected. (d) Fully cured chamber structure incorporating inner chamber, inextensible layer, fibre reinforcement, and outer chamber. (e) Smooth-Sil~960 is poured into the cap mould, and the chamber structure is dipped to form the cap, with an air tube inserted. (f) Completed actuator with Geometry~A.}
    \label{fig:fabrication-A}
\end{figure*}

All actuators were fabricated in platinum-cure silicones from Smooth-On Inc., USA. The actuator body was cast from \emph{Ecoflex~00–50}, while the caps were cast from \emph{Smooth-Sil~960} to provide higher stiffness.
A fibreglass cloth layer was embedded along the flat side as an inextensible sheet, and Kevlar fibre (Kevlar sewing thread T40, Material Metrics Ltd., UK~\cite{KevlarThread}) was wound circumferentially into the silicone wall over the inner chamber and the inextensible layer to provide radial constraint. Varying the fibre angle, pitch, and winding pattern enables tuning of bending performance, and the layouts implemented in this work include SH and DH windings with turn counts ranging from 9 to 100. Minimum silicone cover thicknesses \(\delta_c\) (curved crown) and \(\delta_f\) (flat side) were selected to mitigate the risk of tearing and leakage during demoulding and repeated pressurisation. Early prototypes used small 3D-printed connectors to improve sealing; however, these were later omitted due to frequent leakage caused by bonding failure between materials. All moulds were custom-designed and printed in \emph{Clear~V4} resin using a Form 4 SLA printer (Formlabs, Somerville, MA, USA). The detailed actuator dimensions are listed in Table~\ref{tab:geomA-params}, and fabrication followed a multi-step process as shown in Fig.~\ref{fig:fabrication-A}.

\subsection{Finite Element Modelling}
FEM was used to capture the actuator’s strongly non-linear, large-deformation response. All simulations were performed in \textit{Abaqus/Standard} (Dassault Systèmes, Vélizy-Villacoublay, France) . The silicone bodies were modelled as nearly incompressible hyperelastic materials; fibre reinforcements were modelled as embedded beams tied to the silicone matrix. Unless stated otherwise, a uniform internal pressure was applied to all inner chamber surfaces and ramped monotonically in time.

Ecoflex~00–50 was represented by an isotropic Mooney–Rivlin model with \(C_{10}=0.022\), \(C_{01}=0.001\). Smooth-Sil~960 (cap) used Mooney–Rivlin parameters \(C_{10}=0.7\), \(C_{01}=0.265\) with compressibility \(D_1=1.25\times 10^{-9}\). Clear~V4 tooling inserts (if included) were modelled as linear-elastic, \(E=2.8~\text{GPa}\), \(\nu=0.35\). Dragon~Skin~30 (legacy cap trials) used Mooney–Rivlin \(C_{10}=0.12\), \(C_{01}=0.12\). The flat inextensible layer was approximated by a Yeoh hyperelastic model (\(C_{10}=3.95\)) \cite{Marechal2020}. Kevlar filaments were represented as circular beams of radius \(r_f=0.103~\text{mm}\) (diameter \(0.206~\text{mm}\); taken from the product specification), using linear-elastic properties \(E=40~\text{GPa}\), \(\nu=0.35\). Prior work \cite{modeling} suggests that modelling the fibre radius at half value can reduce stiffness overestimation in some set-ups; trial runs here indicated a negligible effect on bending predictions for the present geometry, so the true radius was retained .

The silicone and fabric layer were meshed with quadratic hybrid tetrahedra (Abaqus element C3D10H) to accommodate near-incompressibility. The fibres were meshed with quadratic beams (B32). Silicone–fabric and inner–outer chamber bodies were merged where co-moulded, and otherwise tied (``Tie'' constraints). Fibres were tied to the silicone along their trajectories. The fibre paths were generated via a Python script adapted from the Soft Robotics Toolkit FEM example \cite{SRT2025}, parametrised by helix type (SH/DH) and turn count.

To match the experiment, an \emph{encastre} (fully fixed) boundary condition was applied to the cap base surface. The air inlet geometry was omitted for simplicity. A spatially uniform pressure \(p(t)\) was applied to the inner chamber surface(s) of each geometry and ramped from \(0\) to \(100~\text{kPa}\) using a smooth amplitude curve. Quasi-static analyses were used with automatic stabilisation where required.

\section{EXPERIMENTS}
\subsection{Abaqus FEM Simulations}
In the FEM simulations, the applied internal pressure was defined to vary proportionally with the simulation time steps in the output files. 
\begin{figure}[t]
    \centering
    \includegraphics[width=0.8\linewidth]{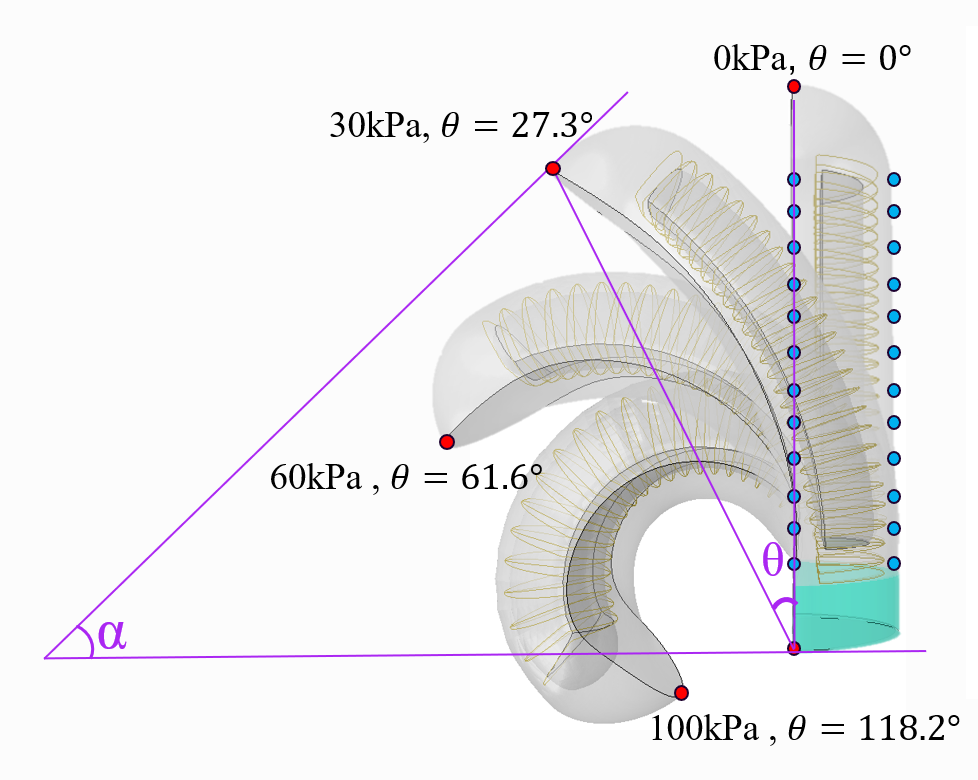}
    \caption{Illustration of the bending angle measurement method used in this study. Representative deformation of 30 DH actuator under FEM increasing internal pressure (0, 30, 60, 100\,kPa). }
    \label{fig:angle_measurement}
\end{figure}
\subsubsection{Bending Angle Simulation}
The corresponding bending angles were computed by tracking the displacement of two terminal nodes, shown as the red dots in Fig.~\ref{fig:angle_measurement}, and calculating the 3D vector angle in \emph{MATLAB}. Specifically, these nodes were identified in the \textit{ABAQUS} model, their history field outputs were requested, and the results were exported to \emph{Excel} with matching time stamps. The time stamps were converted to pressure values, while the node displacements in the \(X\), \(Y\), and \(Z\) directions were combined with their initial coordinates to calculate the bending angle. Tip trajectory plots were also generated for both SH and DH configurations, as well as for different fibre turn counts, based on the displacement-derived bending angle data.

The bending angle in this study is defined as the angle \(\theta\) between the vector from the fixed reference point, the midpoint of the flat-side bottom edge of the actuator cap, and the current position of the actuator tip (Fig.~\ref{fig:angle_measurement}).  Defined angle \(\theta\) is exactly half of the bending angle \(\alpha\) used in several related studies, such as \cite{modeling,Shi2022RoboSoft,Li2023RAL}. By focusing on \(\theta\), the present work simplifies measurement while maintaining consistency between FEM simulation and experimental characterisation.
\subsubsection{Radial Expansion Simulation}
Similarly, radial expansion was measured using a custom \emph{Python} script, which automatically selected 12 pairs of surface nodes on the actuator, as shown by the blue node pairs in Fig.~\ref{fig:angle_measurement}. For each pair, one node was located along the centreline of the flat surface, and the other was positioned on the corresponding point of the opposite curved outer surface. The script identified nodes spaced approximately 2.5\,mm apart along the actuator length, selecting the closest available mesh nodes to the target coordinates. The initial distance between each node pair was taken as the actuator wall thickness; during pressurisation, this distance changed, and the variation was recorded as radial expansion. The script also generated an \emph{Excel} file containing the initial coordinates of all selected nodes, enabling precise calculation of distance changes over time. The displacement–time data were converted into radial expansion–pressure curves, and the results from the 12 node pairs were averaged to obtain the mean radial expansion as a function of pressure.
\subsection{Empirical Experiments}
To experimentally validate the FEM simulations and identify the optimal parameters for the actuator, we tested their bending capability under controlled pressurisation. 
\subsubsection{Experiment Setup}
  
The chamber pressure was regulated via software using an Arduino UNO, which adjusted the voltage supplied to two FESTO VPPX pressure regulators (Esslingen am Neckar, Germany) through a DA4C010BI DAC module (Aptinex, Maharagama, Sri Lanka).  Pressure in the actuator was measured using an ADS1115 16-bit ADC (Adafruit, New York, US) and monitored via the Arduino UNO. The actuator was mounted vertically on an optical table equipped with 3D-printed holders and clamps, ensuring the end cap was rigidly fixed, mimicking the boundary conditions applied in the FEM simulations. A Logitech C920s Pro HD webcam was used to capture image frames for tracking the actuator tip displacement and converting it into bending angles. A checkered calibration board was placed within the field of view to calibrate the images for measurement using the Zhang's method \cite{Zhang2000}. For each applied pressure value, the bending angle relative to the base reference vector is recorded into a CSV file along with annotated images for later analysis. 

\subsubsection{Experiment Protocols}
\paragraph{Bending angle of actuator}
Each actuator was inflated in 10~kPa increments from 0 to 100~kPa, held for 5~seconds per step, then deflated in the same steps. Forward and reverse pressure–angle curves were recorded.
The actuators fabricated for experimental testing were evaluated on the bench; however, due to fabrication constraints including silicone curing, multi-material bonding, and the manual time required for high-turn fibre winding, the range of experimentally produced configurations was more limited than those investigated in the FEM simulations. Furthermore, FEM simulations and experimental protocols were conducted in parallel, allowing findings from each to inform and refine the other.

\paragraph{Bending angle of device}
As the developed actuator in this paper is intended to form part of a larger endoluminal intervention device, we wanted to observe how integration into a larger body affected the bending capability of the actuator. The device in these experiments is a low-fidelity prototype modelled as solid cylinder of diameter 18mm and length 60.5mm, of which 37.5mm is the 30 DH actuator body length. The device is fabricated in EF 00-30 silicone.  Experiments were conducted twice on the same device, first without a camera and then with one embedded, to assess its effect on bending behaviour, as shown in Fig.~\ref{fig:experiment_DH}. Configuration (c) shows the bare device, while (d) shows the identical setup with the camera inserted.

Unlike the actuator-only tests, the device experiments were performed using a syringe to manually inject air from 0–3 ml in 0.2 ml increments, as the device is manually inflated in real-world use.
\section{RESULTS}
\subsection{FEM Simulation Results}
\subsubsection{Actuator Bending Angle}
\begin{figure*}[t]
    \centering
    \includegraphics[width=0.95\textwidth]{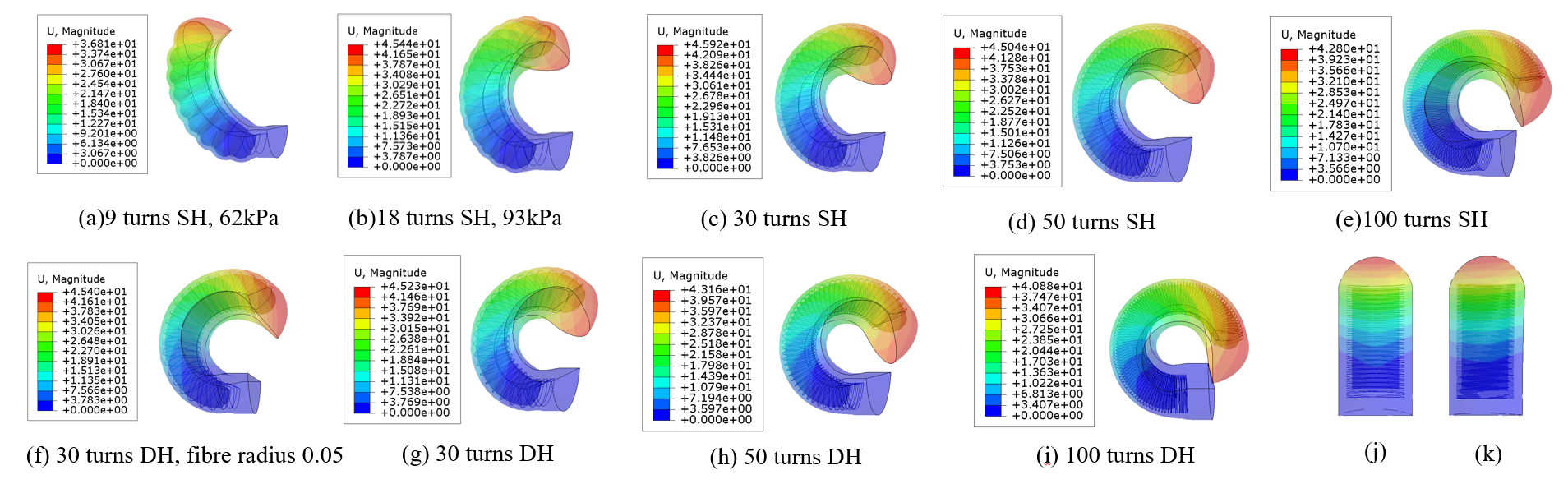}
\caption{FEM simulation results for Geometry~A actuator configurations.  
(a)–(e): SH fibre windings with turn counts of 9, 18, 30, 50, and 100, respectively. Pressures: (a)~62\,kPa, (b)~93\,kPa, (c)–(e)~100\,kPa.  
(f)–(i): DH fibre windings at 100\,kPa. (f)~30 turns with fibre radius halved (0.05\,mm), modelled following~\cite{modeling}. (g)~30 turns, (h)~50 turns, (i)~100 turns DH.  
(j)~DH actuator front view; (k)~SH actuator front view.}
\label{fig:DH_SH_A}

\end{figure*}

Fig.~\ref{fig:DH_SH_A} and and Table~\ref{fig:DH_SH_A} illustrate the fibre configurations simulated during the early stage of actuator development, all based on Geometry~A.  
From Fig.~\ref{fig:DH_SH_A} (a) to (e), the fibre turn count increases from 9 to 100, with all cases employing a SH winding. As the turn count increases, the bulging effect is progressively reduced, radial expansion is increasingly constrained, and the bending angle correspondingly increases. Notably, all SH cases exhibit a degree of asymmetric twisting, as illustrated in (k) at low pressure, whereas the corresponding DH case in Fig.~\ref{fig:DH_SH_A} (j) shows no observable twisting. Quantitatively, the twisting remains minimal, reaching only 2.06\% at 100 kPa as shown in (e). For cases Fig.~\ref{fig:DH_SH_A} (a) and (b), the maximum pressures sustained before simulation failure are 62\,kPa and 93\,kPa, respectively, due to geometric instability leading to simulation abortion. From 30 turns and above, the actuators can withstand pressures exceeding 100\,kPa; hence, these cases are evaluated at the target pressure of 100\,kPa.

Fig.~\ref{fig:DH_SH_A} cases (f), (g), (h), and (i) present DH windings, with fibre turns increasing from 30 to 100, all at 100\,kPa. Similar to the SH results, increasing the number of turns increases the bending angle. Case (f) is identical to case (g) except for the fibre profile: here, the Kevlar fibre radius is modelled as half its actual value (0.05\,mm), following the finite element modelling technique described in~\cite{SRT2025}. However, the effect of this reduced fibre radius on the bending behaviour is negligible compared to case (g).

When comparing SH and DH vertically for the same turn count, DH produces a larger bending angle. This is partly because DH windings have approximately twice the winding density, owing to the forward and backward paths of the same fibre. The largest bending angle is observed in case (i), a DH winding with 100 turns, which, at 100\,kPa, reaches the theoretical maximum bending of $\theta = 180^{\circ}$ (equivalent to $\alpha = 360^{\circ}$ when measured using the notation method defined in Fig.~\ref{fig:angle_measurement}).
\begin{table}[ht]
    \centering
    \caption{Bending angle of configurations shown in Fig.~\ref{fig:DH_SH_A}.}
    \label{tab:DH_SH_A_angles}
    \begin{tabular}{c p{3.5cm} c}
        \hline
        Figure Ref. & Actuator Configuration & Bending Angle $\theta$ ($^\circ$)\\
        \hline
        (a) & 9 turns SH at 62\,kPa & 61.12 \\
        (b) & 18 turns SH at 93\,kPa & 86.12 \\
        (c) & 30 turns SH at 100\,kPa & 90.00 \\
        (d) & 50 turns SH at 100\,kPa & 106.04 \\
        (e) & 100 turns SH at 100\,kPa & 148.80 \\
        (f) & 30 turns DH at 100\,kPa (fibre radius halved) & 98.72 \\
        (g) & 30 turns DH at 100\,kPa & 99.05 \\
        (h) & 50 turns DH at 100\,kPa & 98.13 \\
        (i) & 100 turns DH at 100\,kPa & 180.00 \\
        \hline
    \end{tabular}
\end{table}
\subsubsection{Actuator Radial Expansion}
\begin{figure}[t]
    \centering
    \includegraphics[width=0.95\linewidth]{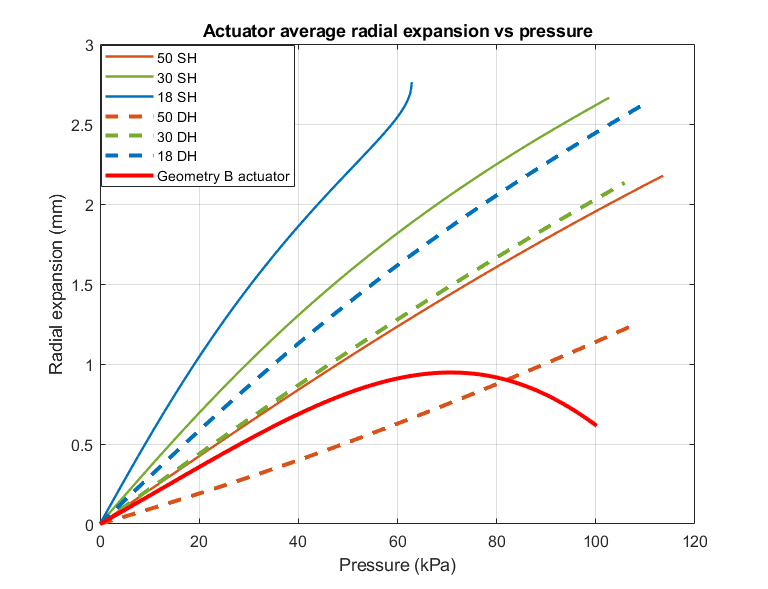}
    \caption{\small 
        Comparison of average radial expansion for selected fibre winding configurations.}
    \label{fig:average_radial}
\end{figure}

In FEM simulation of actuator,  a higher number of fibre turns generally reduces radial expansion, and DH winding produces less radial expansion than SH winding. Fig.~\ref{fig:average_radial} presents the radial expansion curves for several representative configurations. 

The 18-turn SH configuration shows the largest radial expansion, reaching up to 2.6\,mm. Above 60 turns, a sudden increase in slope is observed, likely due to a tracking node being located at a radial expansion peak which, at higher pressures, manifests as a bulge. For later introduced Geometry~B, the radial expansion follows an increase–then–decrease pattern, remaining below 1\,mm across the full pressure range.

\subsubsection{Device Bending Angle}
\begin{figure*}[t]
    \centering
    \includegraphics[width=0.7\textwidth]{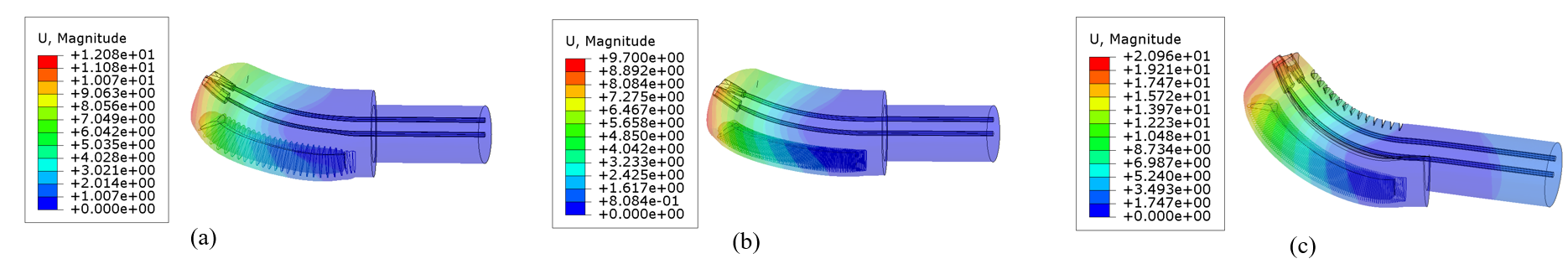}
    \caption{FEM simulation results for the device. (a)~30-turn DH; (b)~100-turn DH. (c) ~100-turn DH with modified side grooves}
    \label{fig:device_A}
\end{figure*}

Fig.~\ref{fig:device_A} presents FEM simulation results of actuators integrated into the complete device assembly. Case~(a) with 30Dh actuator reaches $\theta = 18.1^{\circ}$, case~(b) with a 100 turn DH actuator reaches $\theta = 14.6^{\circ}$, case(c) explores modifications to the 100-turn DH actuator geometry from case~(b) by introducing side-groove designs aimed at enhancing bending performance, which reaches $\theta = 33.52^{\circ}$. All cases in this group are evaluated at 100\,kPa. 

\subsection{Experiment Results}
\subsubsection{Actuator Bending Angle}
\begin{figure}[t]
    \centering
    \includegraphics[width=0.95\linewidth]{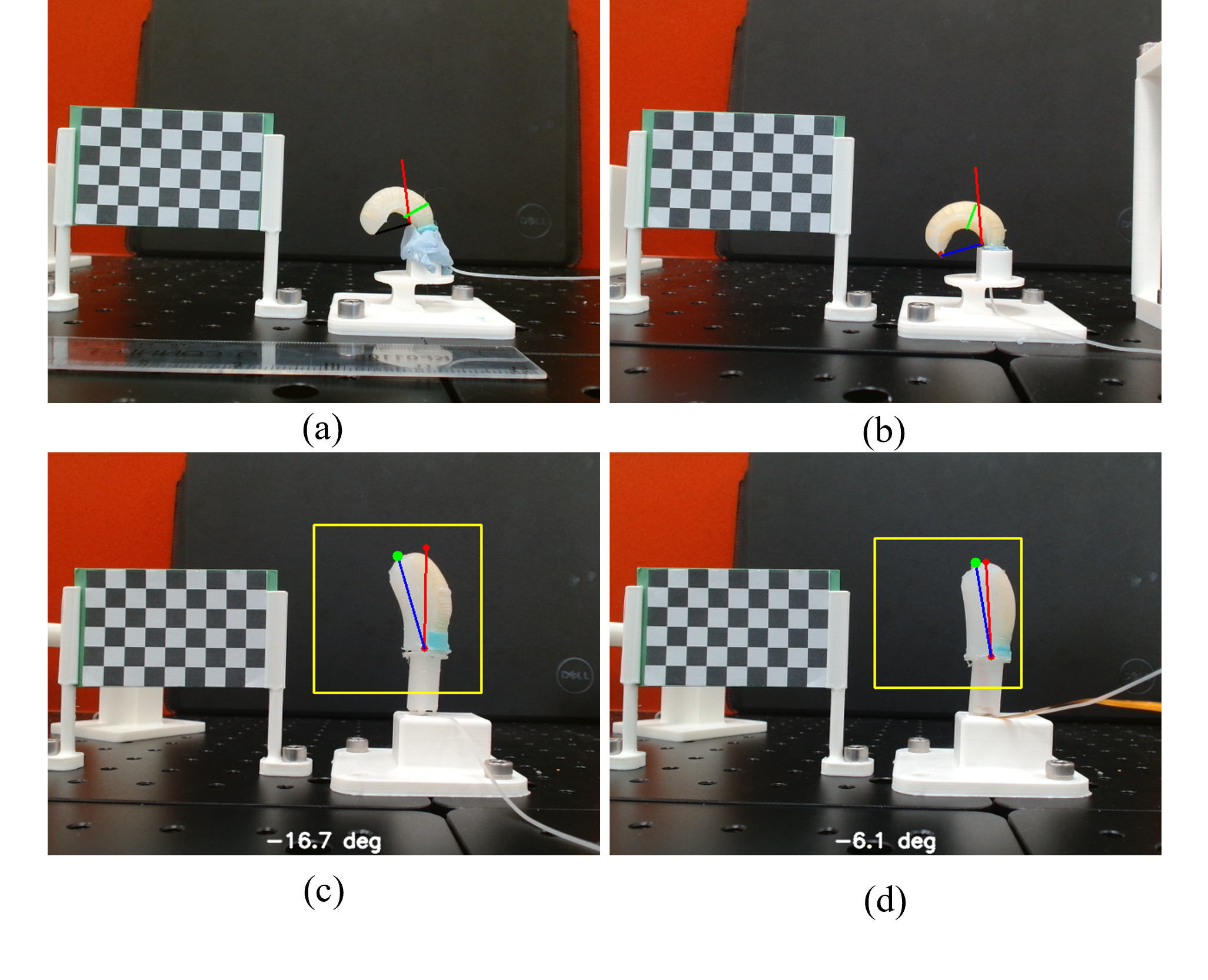}
    \caption{\small 
       Representative Experimental bending angle at 100kPa. (a)~30 DH actuator. (b)~100 SH actuator. (c) 30 DH with the Device. (d) 30 DH with the Device embedded camera.}
    \label{fig:experiment_DH}
\end{figure}

Fig.~\ref{fig:experiment_DH}(a) and (b), showing the maximum bending angle achieved by actuators across multiple experimental trials. Ideally, each actuator would complete up to five trials, with results averaged to provide a representative performance measure. The 30 turns DH actuator reaches $\theta = 85.9^{\circ}$ at 98kPa, The 100 turn SH actuator reaches $\theta = 101.4^{\circ}$ at 100kPa. 

\subsubsection{Device Bending Angle}
Fig.~\ref{fig:experiment_DH}(c) and (d), showing the maximum bending angle achieved by the devices with configuration of Fig.~\ref{fig:device_A} Two devices were tested, in configuration (c), the device consists solely of the silicone structure and the actuator with 30-turn DH winding. In configuration (d), a camera is embedded inside the device, while the remaining structure is identical to that in (c). As expected, the device housing imposes constraints on the actuator, with the thicker structure reducing the achievable bending angle, and the inclusion of the camera further limiting bending. The maximum bending angle observed was $\theta = 16.7^{\circ}$ for (c), compared with a simulated maximum of  $\theta = 18^{\circ}$ at 100~kPa. The bending angle for (d) is  $\theta = 6.1^{\circ}$ .

\subsubsection{Actuator Configuration Choosing}
\begin{figure}[t]
    \centering
    \includegraphics[width=0.9\linewidth]{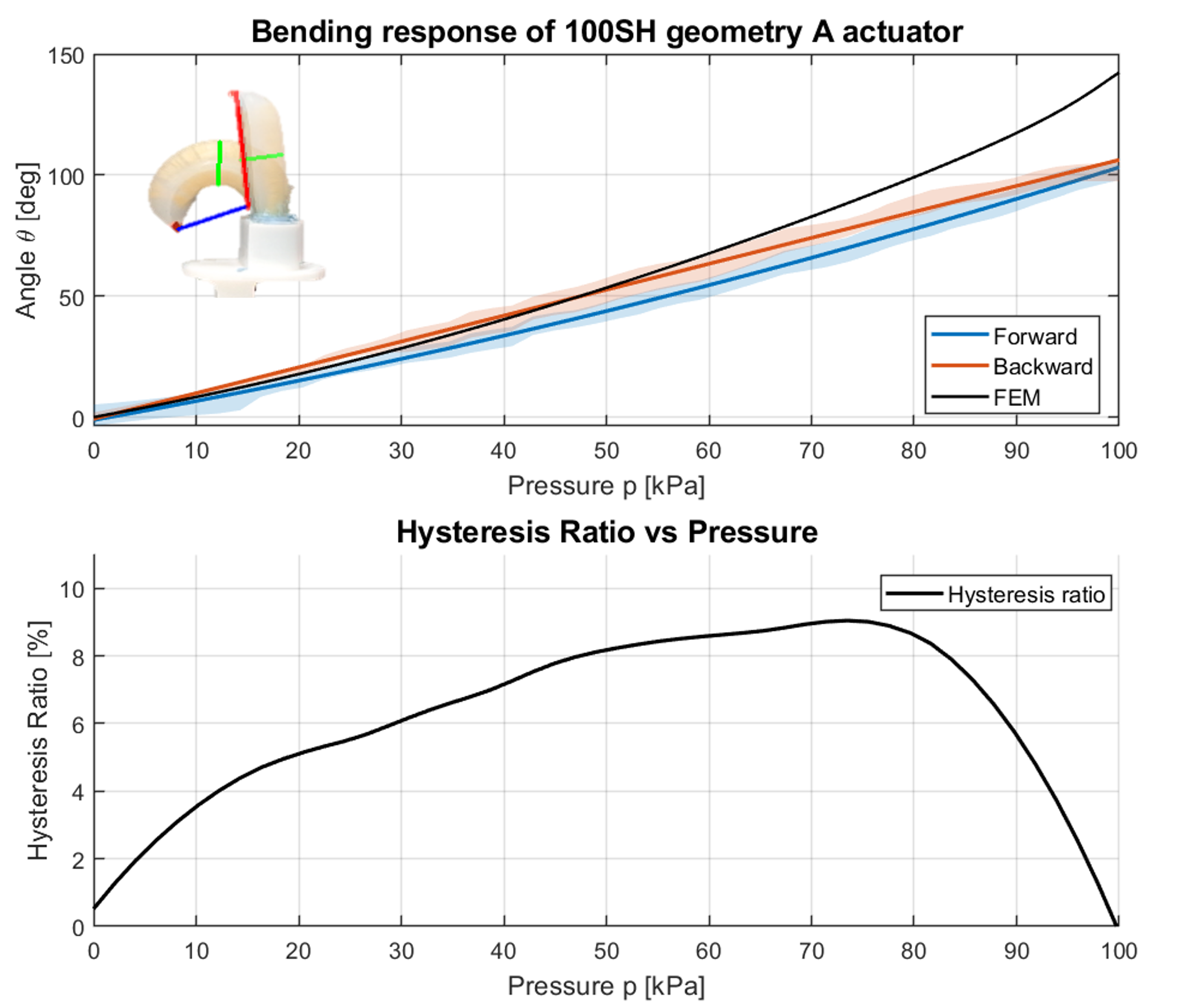}
    \caption{Hysteresis analysis for the 100-turn SH winding actuator: bending angle–pressure curve (forward and backward) and hysteresis ratio–pressure curve.}
    \label{fig:h_ratio}
\end{figure}

The largest experimental bending angle for actuator was achieved with the 100-turn SH winding configuration, as shown in Fig.~\ref{fig:experiment_DH}. This configuration reached a maximum bending angle of  $\theta = 101.4^{\circ}$ at 100\,kPa.  The bending effect and the hysteresis curve, including the forward and backward paths, are presented in Fig.~\ref{fig:h_ratio}.

 From these measurements, the hysteresis ratio was calculated to be 6.28\%.

\section{DISCUSSION}
\subsection{Comparison and Evaluation of FEM and Experiments}
From the simulation results, DH windings generally yielded a greater bending angle compared to SH windings. Nevertheless, in physical testing, the DH configuration introduced a significant practical issue: in every loop, the fibre paths cross, creating concentrated stress points at each intersection. Repeated pressurisation cycles led to localised wear and, ultimately, either fibre breakage or leakage of the actuator. This failure mode was consistently observed across all DH-fabricated actuators.

Leakage analysis for fabricated actuators revealed that most failures occurred at the structural weak points, such as pre-existing air bubbles or imperfect bonding between the chamber and cap. In addition to these visible defects, several actuators appeared well-sealed during initial trials but later developed air bubbles along the outer bending wall, eventually leading to rupture. This behaviour is attributable to detachment between the outer silicone layer and the fibre reinforcement. The thin wall geometry of Geometry~A, with certain interface regions measuring only 0.2–0.3\,mm, is particularly vulnerable. Tiny air pockets, caused by unavoidable dust contamination or stray fibre filaments during fabrication, can accumulate within the pressurised chamber. The Kevlar fibre, being significantly stiffer than the Ecoflex~50 chamber material, can gradually cut into the inner chamber wall under cyclic pressurisation. This process allows air to migrate into the interface between the inner chamber, fibre reinforcement, and outer chamber wall, ultimately creating a structurally weak layer prone to leakage. Once this delamination occurs, the defect is not practically repairable.

As a result, the configuration strategy was shifted towards SH winding. To maintain high bending performance, a 100-turn SH winding was selected for the final fabricated designs from Fig.~\ref{fig:experiment_DH}(a) to (b). From these results, it is evident that the SH configuration exhibits improved structural robustness, with all tested actuators surviving all five trials. Furthermore, the large number of fibre turns provides a favourable bending performance, yielding the best experimental results for the actuator.

The FEM predicts a bending angle of $\theta = 148.8^{\circ}$ at 100\,kPa for 100-turn SH configuration, which is approximately \textbf{38.0\%} higher than the experimental value of $\theta = 101.4^{\circ}$. This difference can be attributed to two main factors: the hyperelastic material model used in Abaqus and variations in the fabrication process. In practice, differences in silicone mixing ratios, sealing quality, actual chamber size, and geometric tolerances mean that each actuator, although fabricated using the same moulds and procedures, will exhibit some variability. In contrast, the Abaqus model is idealised and geometrically perfect. For instance, in Abaqus, all silicone parts are merged to form a perfect bond, which differs significantly from real-world conditions, where silicone layers from different materials or fabrication stages do not always bond perfectly. Furthermore, the hyperelastic model used in Abaqus does not perfectly match the actual material properties of the fabricated actuator. These factors likely contribute to the observed performance gap. Notably, at lower pressures, the difference between FEM predictions and experimental results is negligible, indicating that the FEM simulations are generally consistent with fabrication outcomes in that range.

When compared to similar actuators reported in the literature, the average experimental bending angle of $\theta = 101.4^{\circ}$ (equivalent to $\alpha = 202.9^{\circ}$ at 100,kPa) is highly competitive among miniature actuators ranging from sub-centimetre to a few centimetres-scale. For fairness of comparison, all referenced bending angles were also measured at approximately 100,kPa. In \cite{Shi2022RoboSoft}, a single-chamber actuator achieved 148.12$^\circ$, while the two-chamber configuration reached 222$^\circ$. In \cite{Li2023RAL}, the reported bending angle was approximately 50$^\circ$, and in \cite{modeling} the maximum bending angle across all geometries was around 100$^\circ$. Similarly, \cite{mininature} reported 143.1$^\circ$ for a two-chamber design. Based on these comparisons, our SH 100-turn actuator’s average bending angle of $\alpha = 202.9^{\circ}$ represents a notable improvement over existing designs in this size range.

\begin{figure*}[t]
    \centering
    \includegraphics[width=\textwidth]{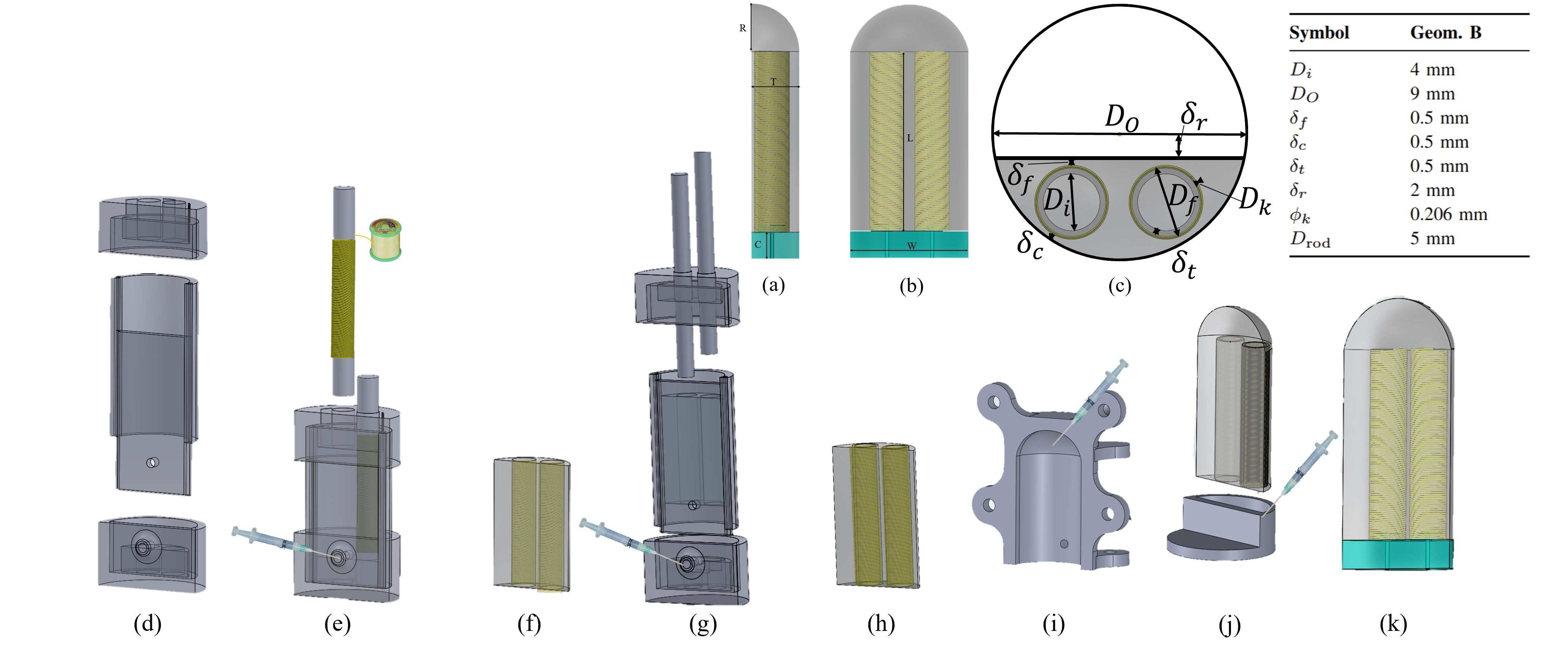}
    \caption{Cross-sectional dimensions and structure of the actuator geometry; Fabrication of actuator with Geometry~B: Symbols correspond to Table. (a) Side view (b) Front view (c) Cross-sectional view. (d) Assembly of outer chamber moulds. (e) Kevlar fibre wound around 5~mm diameter metal rods (one clockwise, one counter-clockwise) are inserted into the assembled outer chamber moulds, and Ecoflex~00–50 is injected. (f) Cured outer chamber with fibre embedded. (g) Assembly of inner chamber moulds with 4~mm diameter rods; Ecoflex~00–50 is injected to form the inner chamber, enclosing fibre between outer and inner layers. (h) Cured tubular chamber structure. (i) Formation of a spherical cap to seal one end. (j) Dipping to form the cap and seal the actuator, with air tube inserted. (k) Completed actuator with Geometry~B.}
    \label{fig:fabrication-B}
\end{figure*}

A challenge encountered in the device experiments was leakage from small, fragile points in the device structure.  At present, a small leakage due to fabrication imperfections is not critical when using a pressure regulator, as the system mechanically maintains constant pressure even if air escapes slowly. However, in real applications, a syringe will be employed to pressurise the actuator. This means that a stable pressure will not be maintained within the chamber.in syringe-based tests, this effect becomes more pronounced, as any leakage prevents the bending effect from being sustained for long durations when sealing is imperfect. This leakage caused the bending angle to decrease over time, meaning that measurements were influenced by the duration of each test; the longer each measurement took, the smaller the recorded angle became. It should be noted that the recorded angles may still be slightly lower than the true maximum achievable angles.

To address this issue, an approach is to enhance the sealing of the actuator by refining the fabrication process or optimising the geometry design. To explore this further we investigated the concept of two parallel circular chambers within the actuator to replace the single semi-circular cavity. This design is referred to as Geometry B (see Fig. \ref{fig:fabrication-B} (a) - (c)). The rationale behind this was that this would enable thicker silicone walls between the cavities and the outer diameter of the actuator, reducing the risk of fabrication defects and related failures. Furthermore, by winding the fibres in opposing directions on parallel chambers, we can eliminate twisting without having to employ DH winding. Another consequence of removing the need for DH winding is ease of fabrication; manually winding around in one direction around a cylinder is notably easier than double winding around a semi-cylinder. The potential of this actuator design to overcome some of the limitations of Geometry A was evaluated using FEM.

\subsection{Enhanced Actuator Design (Geometry B)}

\begin{figure}[t]
    \centering
    \includegraphics[width=0.95\linewidth]{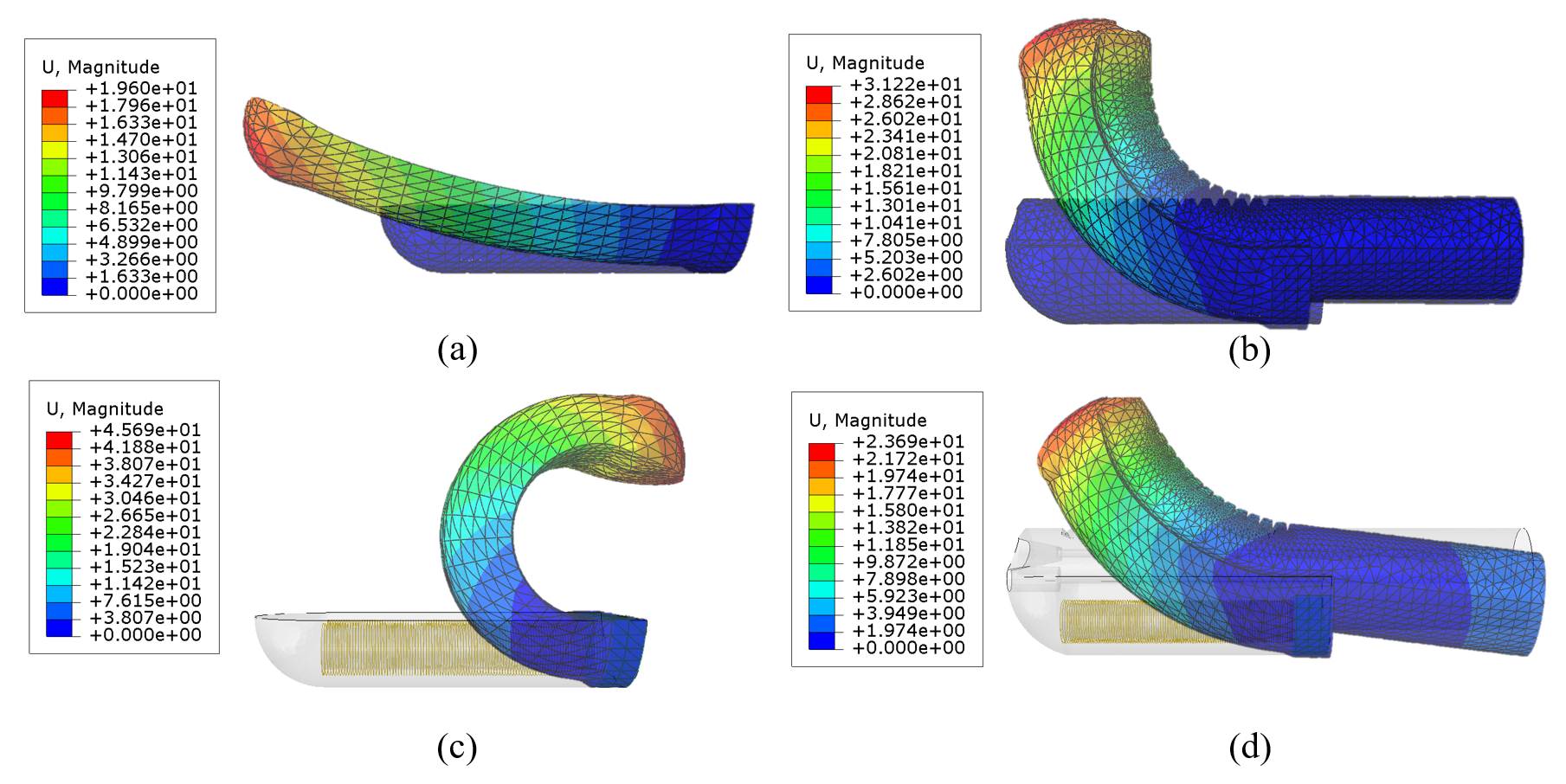}
    \caption{FEM simulation results for Geometry~B under various configurations: (a) actuator alone, elongation-dominated deformation at 100~kPa; (b) actuator integrated into device. (c) actuator with embedded fibreglass inextensible layer. (d) fabric-reinforced actuator integrated into device.}
    \label{fig:geometry_B_FEM}
\end{figure}

The validated settings from Geometry~A (Fig. \ref{fig:cross-sections}) informed the initial configuration of Geometry~B, a double chamber actuator design. The outer envelope is kept identical. The minimum silicone thickness is 0.5~mm for Geometry~B compare to 0.3~mm for Geometry A, the simple chamber structrue above . Chamber depth is 26.5~mm for both designs. The cross-sectional chamber area is  \(25.13~\mathrm{mm}^2\) for Geometry~B, corresponding to a total chamber volumes of approximately  \(666~\mathrm{mm}^3\).

The geometry of Geometry B is shown in Fig.~\ref{fig:fabrication-B}(a)-(c), the two cylindrical chamber were reinforced by one clockwise 100 turn single helix winding and one counter-clock wise 100  turn single helix winding respectively. This symmetric configuration cancelled out the twisting effect introducing by single helix winding. 

Fabrication of the Geometry B actuator required a multi-step process, as show in Fig. \ref{fig:fabrication-B}, which could be a method in future research. Early stage FEM validation of geometry B is been conducted by Abaqus simulation of the actuator and the device using same setting as above sections. 

Fig.~\ref{fig:geometry_B_FEM} presents four sets of FEM simulation results for Geometry~B under different configurations.

In the second case (Fig.~\ref{fig:geometry_B_FEM}b), the actuator is integrated into the complete device assembly. At 95.8~kPa, a bending angle of $\theta = 48.6^{\circ}$ is observed. The increase in bending relative to the actuator alone is primarily due to the thicker outer silicone layer of the device, which acts as a partial constraint on radial expansion.

The third case (Fig.~\ref{fig:geometry_B_FEM}c) introduces an embedded fibreglass inextensible layer identical to the one used in Geometry~A. When pressurised to 100~kPa, the actuator exhibits a pronounced bending of approximately $\theta = 108.6^{\circ}$. The inextensible layer effectively restricts radial expansion, thereby promoting axial elongation and curvature. This configuration demonstrates the significant impact of reinforcement on enhancing bending performance.

Finally, in the fourth case (Fig.~\ref{fig:geometry_B_FEM}d), the fabric-reinforced actuator is integrated into the device. At 97~kPa, the bending angle reaches approximately $\theta = 37.8^{\circ}$, which is lower than the fabric-only configuration. This reduction may be due to the interaction between the inextensible layer and the surrounding device structure, which modifies the deformation path. Plus, the average radial expansion of geometry B actuator is small , with a maximum expansion below 1 mm over the pressure range, as shown in Fig. \ref{fig:average_radial}.

From these observations, it is evident that device implemented geometry B shows a larger bending angle of $\theta = 48.6^{\circ}$ compare to $\theta = 18^{\circ}$in same simulation configuration using single chamber geometry showed in Fig. \ref{fig:device_A}(c). Therefore, Geometry B might be a future exploration direction to improve the bending effect of the device, if the fabrication challenges can be overcome. Another potential future experiment would be to characterise the tip force–pressure relationship of the actuator and, if applicable, to map its workspace.

\section{CONCLUSIONS}
The proposed Miniature soft bending actuator with a 100-turn single-helix configuration demonstrates a large bending capability while retaining a compact centimetre scale form factor. Experimental characterisation achieved bending angles of  $\theta = 101.4^{\circ}$($\alpha = 202.9^{\circ}$), while finite element analysis predicted up to $\theta = 148.8^{\circ}$($\alpha = 297.6^{\circ}$), indicating that the present prototype already approaches the expected upper-bound performance and that further refinement in fabrication accuracy may close the remaining gap. The reasonable correspondence between FEM predictions and the observed deformation trend validates the modelling approach and supports the structural suitability of this configuration for soft continuum robotics at miniature scale. In addition, preliminary simulations suggest that the alternative Geometry B layout could yield even greater bending ranges by improving cross-sectional space utilisation, offering a promising future optimisation pathway for miniature soft endoluminal intervention actuators.


\end{document}